\documentclass{article}

\usepackage[preprint]{tmlr}

\usepackage[utf8]{inputenc}
\usepackage{hyperref}
\usepackage{url}
\usepackage{booktabs}
\usepackage{amsfonts}
\usepackage{amsmath}
\usepackage{amssymb}
\usepackage{nicefrac}
\usepackage{microtype}
\usepackage[table]{xcolor}
\usepackage{graphicx}
\usepackage{subcaption}
\usepackage{pgfplots}
\usepackage{pgfplotstable}
\pgfplotsset{compat=1.18}
\usepgfplotslibrary{colorbrewer,groupplots}
\usetikzlibrary{patterns,positioning}

\definecolor{cbblue}{HTML}{0072B2}
\definecolor{cborange}{HTML}{E69F00}
\definecolor{cbgreen}{HTML}{009E73}
\definecolor{cbred}{HTML}{D55E00}
\definecolor{cbpurple}{HTML}{CC79A7}
\definecolor{cbcyan}{HTML}{56B4E9}
\definecolor{cbyellow}{HTML}{F0E442}
\usepackage{multirow}
\usepackage{makecell}
\usepackage{xspace}
\usepackage{enumitem}

\setlist{nosep,leftmargin=*}

\newcommand{\rankme}{\textsc{RankMe}\xspace}

\title{The Geometric Anatomy of Capability Acquisition in Transformers}
\author{Jayadev Billa\thanks{Unaffiliated researcher - previously at ISI@USC, Yahoo, Nuance, BBN.}\\
San Jose, CA, USA\\
\texttt{jbilla2004@gmail.com}
}

\begin{document}

\maketitle
\setcounter{footnote}{0}

\begin{abstract}

Neural networks gain capabilities during training, but the internal changes that precede capability acquisition are not well understood. In particular, the relationship between geometric change and behavioral change, and the effect of task difficulty and model scale on that relationship, is unclear. We track geometric measures and linear probes across six transformer sizes (405K--151M parameters), eight algorithmic tasks (144 task$\times$level$\times$model combinations), and three Pythia language models (160M--2.8B). Across all settings, representations first collapse to a low-dimensional state, then recover, and only then does behavioral performance improve. Linear probes show that the model's hidden states already contain task-relevant information before the model can act on it.  The collapse floor is task-specific, the collapse propagates top-down through the network, and of the  geometric measures tested, only \rankme reliably precedes capability acquisition for hard tasks. Whether this precursor is detectable depends on task difficulty relative to model capacity. For hard tasks, there is a clear gap: geometry changes first, behavior follows. For easy tasks, the model learns so quickly that both happen simultaneously and no precursor is detectable. On Pythia-2.8B, a logical deduction task that is genuinely hard for the model shows a precursor gap of ${\sim}$49K training steps, while easy benchmarks show none. This suggests that geometric patterns observed in small proxy models can persist at larger scale when the task remains difficult relative to model capacity.

\end{abstract}

\section{Introduction}
\label{sec:introduction}

Neural networks acquire new capabilities during training.  Recent work offers some insight into the internal changes leading to each new capability.  Representation geometry studies have identified distinct developmental stages (collapse, expansion, compression)~\citep{li2025representation}.  Theoretical analysis of attention training has revealed a two-stage process: condensation, followed by rank collapse~\citep{chen2025condensation}.  The grokking literature has identified a transition from memorization to generalization, manifesting as competition between loss basins of different complexity~\citep{nanda2023progress, cullen2026grokking}.  Singular learning theory has introduced the local learning coefficient (LLC) as a proxy for progress~\citep{watanabe2009algebraic, lau2023quantifying, hoogland2024developmental}.  What remains unclear is the temporal relationship between these changes and capability acquisition, and whether their relative ordering holds across tasks and model sizes.  If geometric changes reliably precede capability acquisition, they could provide a basis for monitoring or intervention.

To better understand the temporal relationship between geometric changes and capability acquisition, we create a controlled testbed, consisting of six decoder-only transformer models (405K to 151M parameters), eight algorithmic tasks at three levels of difficulty, and dense checkpointing (206 to 256 per run), resulting in 144 task $\times$ level $\times$ model combinations. At each checkpoint, we measure \rankme, and at select checkpoints, primarily nano-scale, four other geometric measures (gradient effective rank, LLC, Hessian eigenvalues, gradient covariance rank) and train per-task linear probes on the correct output token. We define capability acquisition as the first sustained crossing of 50\% accuracy (remaining above for three consecutive checkpoints). This definition is simply a practical choice and does make a claim about a phase transition. To confirm that patterns hold across larger models scales we test on three Pythia language models (160M, 410M, 2.8B).

A single recurring pattern ties the results together. Consistent with the developmental phases identified by \citet{li2025representation}, representations collapse to a low dimension and then recover, and only then does behavioral accuracy increase. Linear probes confirm that at checkpoints where the model cannot yet perform the task, a trained linear probe on the hidden states can already extract the correct output token. We find that conditioning on individual tasks reveals structure that global analysis does not capture:

\begin{enumerate}[leftmargin=2em]
    \item \emph{The collapse floor is task-specific}
    (\S\ref{sec:collapse-floors}).  Modular arithmetic collapses to  \rankme$\,\approx 2.0$ across a 370$\times$ parameter range (CV\,$=$\,0.16); multiplication floors rise with model capacity (CV\,$=$\,0.35).  The floor appears to reflect the minimum dimensionality the task requires.
  
    \item \emph{The collapse is top-down}
    (\S\ref{sec:layer-propagation}).  Output-facing layers collapse most, early layers least, in all 32/32 task$\times$model combinations tested, consistent with the finding that outer-layer parameters reorganize first~\citep{chen2025condensation}, and not consistent with the assumption that features build bottom-up from simple to complex.  Probing at a layer level confirms that hidden learning is concentrated in the same deep layers where collapse is strongest.

    \item \emph{\rankme is the only reliable precursor}
    (\S\ref{sec:geometric-hierarchy}).  We call a geometric measure a \emph{precursor} if it shows a discrete transition before the model acquires the capability.  Across the task-specific measures tested, \rankme provides the cleanest signal: it precedes capability acquisition for all hard tasks at all tested scales (100\% precursor rate).  Task-specific Hessian and gradient covariance also reach 100\% at nano scale but are too noisy to serve as reliable monitors.  Gradient effective rank transitions too late for most tasks (38\%).  LLC shows no discrete precursor event, consistent with analyses where LLC tracks rather than predicts transitions~\citep{cullen2026grokking}.

    \item \emph{Precursor detectability depends on task difficulty relative to model capacity}
    (\S\ref{sec:geometric-hierarchy}).  For tasks that are hard relative to the model's capacity, \rankme precedes acquisition at every scale tested (100\%).  Easy tasks are acquired during the collapse itself, so no temporal gap exists and no precursor is detectable.  On Pythia-2.8B, a logical deduction task that is genuinely hard for the model (50\% accuracy at step 143K) shows a geometric precursor gap of ${\sim}$49K steps, while easy benchmarks on the same model show none (\S\ref{sec:pythia-validation}).
  \end{enumerate}

Task-specific \rankme tells you whether a particular capability is coming, but not when relative to other tasks.  The geometric dynamics themselves, however, are scale-invariant: collapse floors, phase boundaries, and layer propagation patterns observed at 405K parameters correctly predict the dynamics at 151M and on Pythia across a 17.5$\times$ scale gap ($\rho > 0.92$ between Pythia-160M and 2.8B at 71/72 training steps).  A small proxy model provides the geometric roadmap for a large training run (\S\ref{sec:discussion}).


\section{Related Work}
\label{sec:related-work}

\emph{Representation geometry during training.}
\citet{li2025representation} study the eigenspectra of hidden-state covariance matrices across Pythia training and identify three developmental phases: collapse, expansion and compression. We reproduce this three stage pattern here on Pythia, but focus on a different but related question: do the geometry changes \emph{precede} capability acquisition, which  \citet{li2025representation} do not test. Our task-conditioned study reveals patterns (such as task-dependent floors and top-down propagation) not visible in their global approach. \citet{chen2025condensation} provide a theoretical account in which they show that outer layer parameters reorganize first while attention parameters remain static. The per-layer \rankme analysis presented here is consistent with this outer layer reorganization and further shows that collapse starts at the output layer and propagates inward. \citet{belrose2024statistics} show that models learn statistics of increasing complexity over training, which is also consistent with our observation that representational complexity increases after the initial collapse.

\emph{Loss landscape geometry.} The Hessian eigenspectrum has been shown to develop a bulk-plus-outlier structure during training~\citep{ghorbani2019investigation, sagun2018empirical, papyan2018full}, and the loss barriers between adjacent checkpoints give a complementary view of the landscape connectivity~\citep{vlaar2022linear}. We study both, but find neither gives a signal before capability acquisition (\S\ref{sec:geometric-hierarchy}).

\emph{Singular learning theory.} The LLC $\lambda$ extends the notion of model dimensionality to singular settings~\citep{watanabe2009algebraic}. \citet{lau2023quantifying} derive practical estimators based on SGLD sampling; \citet{hoogland2024developmental} apply LLC to a small transformer trained on language data and find that LLC transitions coincide with behavioral changes. Similarly, \citet{cullen2026grokking}'s analysis of grokking finds that the LLC tracks but does not predict transitions between basins. Nano scale transformer measurements in our multi-task setting mirror these findings: LLC transitions are concurrent with behavioral transitions.

\emph{Capability acquisition and grokking.}
\citet{wei2022emergent} identify capabilities that appear suddenly with increasing training scale. \citet{schaeffer2023emergent} show that sudden emergence can be a metric artifact: accuracy shows a sharp jump where log-probability reveals gradual improvement.  We confirm this divergence in our data, but find that log-probability ``acquisition'' (crossing a midpoint threshold) occurs when accuracy is near zero, so it does not represent genuine capability acquisition.  We define acquisition using accuracy, which has a clear behavioral meaning (\S\ref{sec:acquisition-landscape}). \citet{nanda2023progress} show that mechanistic progress measures (Fourier components, weight norms) predict grokking before generalization. Our \rankme collapse floor of $\approx$2.0 for modular arithmetic is consistent with their finding that two Fourier components suffice, but our geometric measures are coarser: they are able to detect that some reorganization is occurring, but not what kind of circuit is being learned.

\emph{Pythia.} \citet{biderman2023pythia} released a suite of 16 language models with 154 training checkpoints each, making them a rich dataset for probing training dynamics. We study three model sizes: 160M, 410M, and 2.8B. The collapse-recovery pattern replicates, and the geometric dynamics at the 160M scale predict the dynamics at the 2.8B scale ($\rho > 0.92$). Per-task precursors are absent for the easy benchmarks, and present for the hard benchmark we tested (logical deduction), confirming that precursors are viable only when the task is hard relative to the model's capacity.

\section{Methods}
\label{sec:methods}

\subsection{Algorithmic Training Platform}
\label{sec:platform}

We train small transformers on eight algorithmic tasks simultaneously, ranging from token copying to multi-digit multiplication (Table~\ref{tab:tasks}).  Each task has three difficulty levels that scale input size or numerical range, giving 24 task $\times$ level combinations.  Because we control the tasks, we know exactly when the model acquires each capability and can measure what changes geometrically before, during, and after.

\begin{table}[t]
\centering
\caption{Algorithmic tasks and difficulty scaling. All tasks use the format \texttt{TASK input = output}. Difficulty increases with level via input length or numerical range.}
\label{tab:tasks}
\small
\begin{tabular}{l l l l l}
\toprule
Task & Description & L1 & L2 & L3 \\
\midrule
COPY & Copy tokens & 3 tokens & 5 tokens & 8 tokens \\
REV & Reverse tokens & 3 tokens & 5 tokens & 8 tokens \\
CMP & Compare two numbers & 1 digit & 2 digits & 3 digits \\
PAR & Parity of bits & 4 bits & 6 bits & 8 bits \\
ADD & Addition & 1+1 digit & 2+2 digits & 3+3 digits \\
MOD & Modular arithmetic & $p \in \{2,3,5,7\}$ & $p \in \{7,11,13\}$ & $p \in \{13,17,19,23\}$ \\
SORT & Sort numbers & 3 numbers & 5 numbers & 8 numbers \\
MUL & Multiplication & $1{\times}1$ digit & $1{\times}2$ digits & $2{\times}2$ digits \\
\bottomrule
\end{tabular}
\end{table}

\emph{Architecture.} GPT-2-style decoder-only transformers~\citep{radford2019language} with pre-norm, GELU, learned positional embeddings, tied embeddings, no dropout. Six sizes span 370$\times$ in parameter count (Table~\ref{tab:models}).  The largest (XLarge, 151M) is deliberately sized to match Pythia-160M, bridging the gap between our controlled setting and naturalistic pre-training. 

\begin{table}[t]
\centering
\caption{Model configurations. Parameter counts exclude tied output embedding weights.}
\label{tab:models}
\small
\begin{tabular}{l r r r r r}
\toprule
Size & Layers & $d_\text{model}$ & Heads & $d_\text{ff}$ & Parameters \\
\midrule
Nano & 2 & 128 & 4 & 512 & 405K \\
Micro & 4 & 192 & 6 & 768 & 1.8M \\
Small & 6 & 320 & 8 & 1{,}280 & 7.4M \\
Medium & 8 & 512 & 8 & 2{,}048 & 25.2M \\
Large & 12 & 768 & 12 & 3{,}072 & 85M \\
XLarge & 12 & 1{,}024 & 16 & 4{,}096 & 151M \\
\bottomrule
\end{tabular}
\end{table}

\emph{Training.} AdamW with cosine LR schedule (1K warmup, peak $3{\times}10^{-4}$ for nano--small, $1{\times}10^{-4}$ for medium--xlarge), weight decay 0.1, batch size 64.  Data is generated on the fly with uniform sampling across tasks and levels.  Loss is masked to answer tokens only.  Training runs 100K steps for nano--small, 200K for medium--xlarge.  Character-level tokenization with vocabulary size 41.

Checkpoints are saved densely early (every 100 steps for the first 10K, coarser after that), yielding 206 checkpoints for nano--small and 256 for medium--xlarge (117 for xlarge, which trained for fewer steps).  At each checkpoint we evaluate all 24 task$\times$level combinations on 1{,}000 examples, recording accuracy and log-probability~\citep{schaeffer2023emergent}.

\emph{Acquisition detection.} We define capability acquisition as the first training step at which accuracy $\geq 0.5$ for $\geq$3 consecutive checkpoints.  All precursor statistics for the algorithmic tasks use the accuracy-based definition.  Robustness to threshold choice is verified in Appendix~\ref{app:robustness}.  For Pythia, we use the log-probability analogue since accuracy is not always well-defined on those benchmarks.

\subsection{Geometric Measures}
\label{sec:measures}

We measure five complementary geometric properties spanning representation geometry, gradient geometry, and loss landscape geometry (Table~\ref{tab:measures}). All five are computed at nano scale to establish the temporal hierarchy. RankMe, the computationally cheapest (${\sim}$1\,s/checkpoint at nano scale), is the only measure computed at all six scales; the cross-scale analysis relies on it alone. Gradient effective rank is computed at nano (206 checkpoints) and sparsely at large (37 checkpoints); Hessian and gradient covariance are computed at nano and on Pythia 160M/410M; LLC is computed at nano only.

\begin{table}[t]
\centering
\caption{Geometric measures. Cost is wall-clock time per checkpoint on nano (405K params).}
\label{tab:measures}
\small
\begin{tabular}{l l l l}
\toprule
Measure & Captures & Formulation & Cost \\
\midrule
\rankme & Repr.\ dimensionality & $\exp(H(\bar{\sigma}))$, $\bar{\sigma}_i = \sigma_i / \sum_j \sigma_j$ & $\sim$1s \\
Grad.\ eff.\ rank & Gradient concentration & $\exp(H(\bar{\lambda}))$ of gradient eigenvalues & $\sim$85s \\
LLC & Loss landscape complexity & SGLD loss elevation $\hat{\lambda}$ & $\sim$11s \\
Hessian top-$\lambda$ & Curvature & Stochastic Lanczos, $k{=}5$ (per task) & $\sim$1s \\
Grad.\ cov.\ rank & Gradient diversity & Effective rank of $\nabla L \nabla L^\top$ (per task) & $\sim$6s \\
\bottomrule
\end{tabular}
\end{table}

\emph{Task-specific} \rankme.  For each task, we collect hidden-state activations from the final transformer layer on 200 diagnostic examples, compute the singular value decomposition, and apply the \rankme formula~\citep{garrido2023rankme}: $\text{RankMe}(H) = \exp\bigl(-\sum_i \bar{\sigma}_i \log \bar{\sigma}_i\bigr)$, where $\bar{\sigma}_i = \sigma_i / \sum_j \sigma_j$ are the normalized singular values.  This measures the effective dimensionality of the representation space as seen by each task.  We also compute \rankme per transformer layer to study propagation patterns.

\emph{Gradient effective rank.} We compute per-sample gradients on 200 task-specific examples, forming a gradient matrix $G \in \mathbb{R}^{N \times P}$ ($N$ samples, $P$ parameters), and eigendecompose the Gram matrix $G G^\top \in \mathbb{R}^{N \times N}$.  The effective rank is $\exp(H(\bar{\lambda}))$, where $\bar{\lambda}$ are the normalized eigenvalues.  This measures how concentrated the per-sample gradient directions are: low rank means gradients point in similar directions across examples, high rank means they are diverse.

\emph{Local Learning Coefficient (LLC).} Following \citet{lau2023quantifying}, we estimate the LLC at each checkpoint using SGLD sampling: we run 500 SGLD steps with learning rate $10^{-5}$, inverse temperature $\beta = 1.0$, and localization strength $\gamma = 10{,}000$, discarding the first 100 steps as burn-in and computing the mean loss elevation from the remaining 400 steps.  The LLC measures the effective complexity of the loss landscape near the current parameter configuration.

\emph{Hessian top eigenvalue.} We compute the top eigenvalues of the Hessian via stochastic Lanczos quadrature~\citep{ghorbani2019investigation}.  For Pythia, we use random Pile validation data (global, $k{=}20$ Lanczos vectors, ${\sim}$30s/checkpoint).  For the algorithmic experiment, we compute a task-specific variant: Lanczos on each task's loss separately ($k{=}5$, ${\sim}$1s/checkpoint per task).

\emph{Gradient covariance rank.} The effective rank of the batch-level gradient covariance matrix, measuring the diversity of gradient directions across training examples.  Computed per-task on the same diagnostic sets used for \rankme.  For computational tractability, we use the first 50K parameters, which biases toward embedding weights and early layers (see Appendix~\ref{app:implementation} for details).

\subsection{Output Token Probing}
\label{sec:probing-method}

To test whether the geometric reorganization is task-relevant, we train per-task linear probes at each checkpoint.  For each task, we extract final-layer hidden states at answer positions and train a logistic regression to predict the correct next token.  If the probe can extract the correct answer at a checkpoint where the model cannot yet produce it, the representations contain task-relevant information before the model can act on it.

We use 80/20 stratified train/test splits with convergence at 500 iterations.  Per-layer probing applies the same method to each transformer layer's output.  All probing uses the same 200-example diagnostic sets used for \rankme.

\subsection{Pythia Validation}
\label{sec:pythia-setup}

We analyze three Pythia-deduped models~\citep{biderman2023pythia}: 160M (154 checkpoints), 410M (154 checkpoints), and 2.8B (72 checkpoints spanning the full training trajectory).  For each checkpoint we compute task-specific \rankme on seven diagnostic sets (Table~\ref{tab:pythia-diagnostics}).  Unlike the algorithmic experiment, these are external probes, not inputs from the training distribution.

\begin{table}[t]
\centering
\caption{Pythia diagnostic sets. Unlike the algorithmic experiment, these are external probes, not inputs from the training distribution.}
\label{tab:pythia-diagnostics}
\small
\begin{tabular}{l l l r}
\toprule
Benchmark & Capability & Prompt format (example) & $N$ \\
\midrule
Syntactic & Subject--verb agreement & ``The keys to the cabinet \textit{is/are}'' (logprob) & 36 \\
Semantic & Word analogies & ``man : woman :: king : \textit{\_\_\_}'' (4-way choice) & 30 \\
Arithmetic & Few-shot addition/mult.\ & ``3+7=10; 15+23=38; 42+8=\textit{?}'' (3-shot) & 200 \\
ICL & In-context learning & Novel label mappings (4-shot sent./categ.) & 16 \\
Factual & Factual recall & ``The capital of France is \textit{\_\_\_}'' (next token) & 51 \\
Logical & Deductive reasoning & ``All dogs are animals. Rex is a dog. $\therefore$ \textit{\_\_\_}'' & 26 \\
Pile & General text modeling & Random Pile validation sequences (perplexity) & 200 \\
\bottomrule
\end{tabular}
\end{table}

We additionally use global geometric measures (Hessian eigenvalues, loss barriers, gradient covariance) at all 154 checkpoints for 160M and 410M.

\subsection{Analysis Methods}
\label{sec:analysis-methods}

\emph{Temporal precedence.} A geometric measure is a \emph{precursor} for a given task if its characteristic transition (e.g., the \rankme collapse minimum) occurs before the task is acquired.  The precursor rate is the fraction of task$\times$model combinations where the geometric transition precedes acquisition.

\emph{State-based prediction.} We test whether the \rankme value at the collapse floor predicts the order in which tasks are acquired.  We compute Spearman correlations between floor values and acquisition steps, then check whether any predictive power persists within a difficulty class (e.g., does a lower floor among hard tasks predict earlier acquisition?) or is simply driven by the fact that hard and easy tasks have different floors.

\section{Results}
\label{sec:results}

\subsection{The Acquisition Landscape}
\label{sec:acquisition-landscape}

Of 144 task$\times$level$\times$model combinations, 142 achieve sustained acquisition (accuracy $\geq 0.5$ for $\geq$3 consecutive checkpoints).  The two exceptions are MUL\_L3 on nano, which never crosses 50\% accuracy, and MUL\_L3 on xlarge, where training was stopped at 18K steps, before the expected acquisition window (43K--56K at other scales), due to compute constraints.  Since   difficulty levels of any task are related (if a task is hard at L1, it is hard at L2 and L3), the effective number of independent observations is closer to 48 (8 tasks $\times$ 6 scales). That said, the range is large: acquisition spans nearly three orders of magnitude, from step~100 (CMP on xlarge) to step~56{,}000 (MUL\_L3 on small).

We split tasks into two categories based on timing relative to early training dynamics (Table~\ref{tab:emergence-map}). These are static labels which do not reflect the relative task difficulty vis-\`a-vis model capacity. In general, tasks become easier for the model as its capacity increases.  \emph{Easy tasks} (COPY, REV, CMP, PAR, SORT) are typically acquired within the first 2{,}500 steps, during or shortly after the initial representation reorganization; \emph{hard tasks} (ADD, MOD, MUL) take thousands to tens of thousands of steps, with larger models generally acquiring them earlier (e.g., MOD\_L3: 16K steps at nano vs.\ 6.0K at large; full table in Appendix~\ref{app:emergence-tables}).\footnote{SORT, despite involving a multi-element operation, is acquired in the easy range (1{,}500--1{,}800 steps at L3). PAR shows scale sensitivity: PAR\_L2 on small is acquired at step 3{,}700 (hard-task range), and a few other PAR combinations exceed 2{,}500 steps, though the majority are acquired early.}

For every task, accuracy-based and log-probability-based detection diverge: the ``emergence mirage'' of \citet{schaeffer2023emergent}, replicated here in a controlled setting.  Log-probability crosses its midpoint threshold 3--15$\times$ earlier than accuracy crosses 50\%, but at that point accuracy is typically 0--15\%: the model cannot yet perform the task.  We therefore define acquisition using accuracy, which has a clear behavioral interpretation (the model can do the task more often than not).  The full divergence analysis is in Appendix~\ref{app:log-prob-details}.

\begin{table}[t]
\centering
\caption{Acquisition steps (training step at which accuracy $\geq 0.5$ for $\geq$3 consecutive checkpoints) for 8 tasks across 6 model sizes. Easy tasks (gray) are acquired during initialization; hard tasks (white) show scale-dependent acceleration. At xlarge, even hard tasks are acquired within the first few thousand steps. L2 difficulty shown; full tables in Appendix~\ref{app:emergence-tables}.}
\label{tab:emergence-map}
\small
\begin{tabular}{l r r r r r r}
\toprule
Task & Nano & Micro & Small & Medium & Large & XLarge \\
 & (405K) & (1.8M) & (7.4M) & (25.2M) & (85M) & (151M) \\
\midrule
\rowcolor{gray!15} COPY & 700 & 700 & 400 & 600 & 500 & 400 \\
\rowcolor{gray!15} REV & 700 & 600 & 400 & 600 & 400 & 400 \\
\rowcolor{gray!15} CMP & 400 & 400 & 300 & 100 & 200 & 100 \\
\rowcolor{gray!15} PAR & 900 & 900 & 3{,}700 & 2{,}500 & 600 & 900 \\
\rowcolor{gray!15} SORT & 1{,}100 & 900 & 900 & 700 & 900 & 500 \\
ADD & 5{,}700 & 3{,}100 & 3{,}900 & 3{,}200 & 3{,}500 & 2{,}600 \\
MOD & 6{,}500 & 5{,}100 & 5{,}000 & 3{,}600 & 3{,}600 & 3{,}500 \\
MUL & 6{,}900 & 5{,}400 & 4{,}700 & 4{,}100 & 3{,}500 & 3{,}400 \\
\midrule
\multicolumn{7}{l}{\footnotesize XLarge: 23/24 combinations acquired by step 13K; MUL\_L3 not acquired (training stopped at 18K).} \\
\bottomrule
\end{tabular}
\end{table}

\begin{figure}[t]
    \centering
\begin{tikzpicture}
\begin{axis}[
    width=0.7\textwidth,
    height=5.5cm,
    colormap={viridisR}{
        rgb255=(253,231,37)
        rgb255=(94,201,98)
        rgb255=(33,145,140)
        rgb255=(59,82,139)
        rgb255=(68,1,84)
    },
    colorbar,
    colorbar style={
        ylabel={\small Emergence step},
        ytick={2.3,2.7,3,3.7,4,4.7},
        yticklabels={\scriptsize 200,\scriptsize 500,\scriptsize 1K,\scriptsize 5K,\scriptsize 10K,\scriptsize 50K},
        ylabel style={font=\small},
    },
    point meta min=2.2,
    point meta max=4.8,
    xtick={0,1,2,3,4},
    xticklabels={\scriptsize 405K,\scriptsize 1.8M,\scriptsize 7.4M,\scriptsize 25M,\scriptsize 85M},
    ytick={0,1,2,3,4,5,6,7},
    yticklabels={\scriptsize COPY,\scriptsize REV,\scriptsize CMP,\scriptsize PAR,\scriptsize ADD,\scriptsize MOD,\scriptsize SORT,\scriptsize MUL},
    xlabel={\small Model size},
    tick label style={font=\scriptsize},
    label style={font=\small},
    y dir=reverse,
    enlargelimits=false,
    axis on top,
]
\addplot[matrix plot*, mesh/cols=5, point meta=explicit] coordinates {
    (0,0) [2.30]  (1,0) [2.30]  (2,0) [2.30]  (3,0) [2.30]  (4,0) [2.30]
    (0,1) [2.60]  (1,1) [2.48]  (2,1) [2.48]  (3,1) [2.30]  (4,1) [2.30]
    (0,2) [2.60]  (1,2) [2.48]  (2,2) [2.30]  (3,2) [2.30]  (4,2) [2.30]
    (0,3) [2.78]  (1,3) [2.60]  (2,3) [2.48]  (3,3) [2.48]  (4,3) [2.30]
    (0,4) [3.78]  (1,4) [3.93]  (2,4) [3.85]  (3,4) [3.78]  (4,4) [3.74]
    (0,5) [4.20]  (1,5) [4.08]  (2,5) [3.94]  (3,5) [3.72]  (4,5) [3.65]
    (0,6) [3.95]  (1,6) [3.88]  (2,6) [3.81]  (3,6) [3.70]  (4,6) [3.60]
    (0,7) [4.65]  (1,7) [4.75]  (2,7) [4.68]  (3,7) [4.65]  (4,7) [4.60]
};

\draw[white, line width=2pt] (axis cs:-0.5,3.5) -- (axis cs:4.5,3.5);

\end{axis}
\end{tikzpicture}
    \caption{Acquisition map. Log-scaled acquisition step for 8 tasks across 6 model sizes. Easy tasks (above white line) are acquired uniformly early; hard tasks (below) show scale-dependent acceleration. At xlarge (151M), the gap between easy and hard nearly vanishes.}
    \label{fig:emergence-map}
\end{figure}

\subsection{Universal Collapse to Task-Specific Floors}
\label{sec:collapse-floors}

During the first 200--1{,}000 training steps, task-specific \rankme drops sharply across all tasks, model sizes, and layers.  Across all tasks and model sizes we test, the collapse is consistent, and in line with \citet{li2025representation}: all 8 tasks across all 6 model sizes collapse in the same step range with near-identical dynamics.  What differs between tasks is the floor (Figure~\ref{fig:collapse-floors}). MOD collapses to \rankme$\,\approx 2.0$ regardless of model size, consistent with its two-dimensional Fourier structure~\citep{nanda2023progress}.  MUL floors increase with model capacity.

\begin{figure}[t]
    \centering
\begin{tikzpicture}
\begin{groupplot}[
    group style={
        group size=4 by 1,
        horizontal sep=0.4cm,
        y descriptions at=edge left,
    },
    width=0.27\textwidth,
    height=4.5cm,
    xlabel={\small Step},
    xmin=0, xmax=25000,
    scaled x ticks=false,
    xtick={0,12500,25000},
    xticklabels={\scriptsize 0,\scriptsize 12.5K,\scriptsize 25K},
    tick label style={font=\tiny},
    label style={font=\small},
    title style={font=\small\bfseries},
    every axis plot/.append style={line width=0.7pt, mark=none},
]

\nextgroupplot[title={MOD}, ylabel={\small RankMe}, ymin=0, ymax=28,
    legend to name=collapseleg,
    legend style={font=\small, draw=none, legend columns=5, /tikz/every even column/.append style={column sep=6pt}},
]
\fill[gray!10] (axis cs:0,0) rectangle (axis cs:200,28);
\addplot[cbblue, solid] table[x=step, y=rankme, col sep=comma]{data/fig1_mod_nano.csv};
\addlegendentry{405K}
\addplot[cborange, dashed] table[x=step, y=rankme, col sep=comma]{data/fig1_mod_micro.csv};
\addlegendentry{1.8M}
\addplot[cbgreen, dotted, line width=0.9pt] table[x=step, y=rankme, col sep=comma]{data/fig1_mod_small.csv};
\addlegendentry{7.4M}
\addplot[cbred, dashdotted] table[x=step, y=rankme, col sep=comma]{data/fig1_mod_medium.csv};
\addlegendentry{25M}
\addplot[cbpurple, densely dashed] table[x=step, y=rankme, col sep=comma]{data/fig1_mod_large.csv};
\addlegendentry{85M}

\nextgroupplot[title={ADD}, ymin=0, ymax=28]
\fill[gray!10] (axis cs:0,0) rectangle (axis cs:200,28);
\addplot[cbblue, solid] table[x=step, y=rankme, col sep=comma]{data/fig1_add_nano.csv};
\addplot[cborange, dashed] table[x=step, y=rankme, col sep=comma]{data/fig1_add_micro.csv};
\addplot[cbgreen, dotted, line width=0.9pt] table[x=step, y=rankme, col sep=comma]{data/fig1_add_small.csv};
\addplot[cbred, dashdotted] table[x=step, y=rankme, col sep=comma]{data/fig1_add_medium.csv};
\addplot[cbpurple, densely dashed] table[x=step, y=rankme, col sep=comma]{data/fig1_add_large.csv};

\nextgroupplot[title={SORT}, ymin=0, ymax=28]
\fill[gray!10] (axis cs:0,0) rectangle (axis cs:200,28);
\addplot[cbblue, solid] table[x=step, y=rankme, col sep=comma]{data/fig1_sort_nano.csv};
\addplot[cborange, dashed] table[x=step, y=rankme, col sep=comma]{data/fig1_sort_micro.csv};
\addplot[cbgreen, dotted, line width=0.9pt] table[x=step, y=rankme, col sep=comma]{data/fig1_sort_small.csv};
\addplot[cbred, dashdotted] table[x=step, y=rankme, col sep=comma]{data/fig1_sort_medium.csv};
\addplot[cbpurple, densely dashed] table[x=step, y=rankme, col sep=comma]{data/fig1_sort_large.csv};

\nextgroupplot[title={MUL}, ymin=0, ymax=28]
\fill[gray!10] (axis cs:0,0) rectangle (axis cs:200,28);
\addplot[cbblue, solid] table[x=step, y=rankme, col sep=comma]{data/fig1_mul_nano.csv};
\addplot[cborange, dashed] table[x=step, y=rankme, col sep=comma]{data/fig1_mul_micro.csv};
\addplot[cbgreen, dotted, line width=0.9pt] table[x=step, y=rankme, col sep=comma]{data/fig1_mul_small.csv};
\addplot[cbred, dashdotted] table[x=step, y=rankme, col sep=comma]{data/fig1_mul_medium.csv};
\addplot[cbpurple, densely dashed] table[x=step, y=rankme, col sep=comma]{data/fig1_mul_large.csv};

\end{groupplot}

\node[anchor=north] at ([yshift=-0.8cm]$(group c2r1.south)!0.5!(group c3r1.south)$) {\ref{collapseleg}};
\end{tikzpicture}
    \caption{Task-specific collapse floors. \rankme trajectories for three hard tasks across six model sizes. All models collapse to task-specific minima during initialization (shaded), then recover. MOD floor is scale-invariant (${\approx}2.0$, CV = 0.16); MUL floors increase with capacity.}
    \label{fig:collapse-floors}
\end{figure}

\subsection{Top-Down Layer Propagation}
\label{sec:layer-propagation}

Per-layer \rankme reveals that the collapse propagates top-down: the deepest (output-facing) layers collapse most, while early layers retain more representational diversity.  This holds in 32 of 32 task$\times$model combinations tested (8 tasks $\times$ 4 model sizes with $\geq$4 layers: micro through large; xlarge was excluded because per-layer \rankme was not computed at that scale), which is not consistent with the intuition that learned features build bottom-up from simple to complex.

At the collapse minimum, the final transformer layer has much lower \rankme than the first.  For MOD: micro layer~0 = 8.2 vs.\ layer~3 = 1.7 (80\% reduction), small layer~0 = 10.4 vs.\ layer~5 = 2.2 (78\%), medium layer~0 = 14.8 vs.\ layer~7 = 2.3 (84\%).  The gradient is monotonic (layer $L$ collapses more than layer $L{-}1$), and first-to-last-layer reduction ranges from 30\% to 84\% across all tasks and sizes.

After acquisition, the pattern changes: the final layer remains compressed while middle layers recover and diversify (Figure~\ref{fig:layer-propagation}).  During collapse, the output layer commits to a low-dimensional representation; after acquisition, intermediate layers reorganize to support it.

\begin{figure}[t]
    \centering
\begin{tikzpicture}
\begin{groupplot}[
    group style={
        group size=2 by 1,
        horizontal sep=1.2cm,
    },
    width=0.45\textwidth,
    height=4.5cm,
    xlabel={\small Layer index},
    ylabel={\small RankMe},
    tick label style={font=\scriptsize},
    label style={font=\small},
    title style={font=\small\bfseries},
    every axis plot/.append style={line width=0.9pt},
    ymin=0,
]

\nextgroupplot[title={At collapse minimum}, xmin=-0.5, xmax=7.5, xtick={0,1,...,7},
    legend to name=layerleg,
    legend style={font=\small, draw=none, legend columns=3, /tikz/every even column/.append style={column sep=6pt}},
]
\addplot[cborange, mark=*, mark size=2pt] table[x=layer, y=rankme, col sep=comma]{data/fig3_micro_collapse.csv};
\addlegendentry{1.8M (4L)}
\addplot[cbgreen, mark=square*, mark size=2pt] table[x=layer, y=rankme, col sep=comma]{data/fig3_small_collapse.csv};
\addlegendentry{7.4M (6L)}
\addplot[cbred, mark=triangle*, mark size=2.5pt] table[x=layer, y=rankme, col sep=comma]{data/fig3_medium_collapse.csv};
\addlegendentry{25M (8L)}

\nextgroupplot[title={Post-emergence (+10K steps)}, xmin=-0.5, xmax=7.5, xtick={0,1,...,7}]
\addplot[cborange, mark=*, mark size=2pt] table[x=layer, y=rankme, col sep=comma]{data/fig3_micro_post_emergence.csv};
\addplot[cbgreen, mark=square*, mark size=2pt] table[x=layer, y=rankme, col sep=comma]{data/fig3_small_post_emergence.csv};
\addplot[cbred, mark=triangle*, mark size=2.5pt] table[x=layer, y=rankme, col sep=comma]{data/fig3_medium_post_emergence.csv};

\end{groupplot}

\node[anchor=north] at ([yshift=-0.8cm]$(group c1r1.south)!0.5!(group c2r1.south)$) {\ref{layerleg}};
\end{tikzpicture}
    \caption{Top-down layer propagation. Per-layer \rankme at the collapse minimum (left) and post-acquisition (right) for three model sizes on MOD. At collapse, deeper layers show lower \rankme. After acquisition, the final layer stays compressed while middle layers recover.}
    \label{fig:layer-propagation}
\end{figure}

\subsection{Hidden Learning Before Acquisition}
\label{sec:hidden-learning}

The geometric analyses above show that representations reorganize before acquisition, but not whether this reorganization is task-relevant.  We test directly: at each checkpoint, we train a linear probe (logistic regression) on each task's hidden representations to predict the correct output token.

For all 8 tasks across all model sizes, the probe can extract the correct output token at checkpoints where the model cannot yet produce it (Table~\ref{tab:hidden-learning}).  The probe is a trained classifier with its own capacity to learn the mapping from hidden states to output tokens; its success indicates that task-relevant information is present in the representations, not that the model can use it.  Per-layer probing at nano scale shows that the improvement concentrates in the deeper layers (deep-to-shallow $\Delta$ ratio 3--11$\times$ for ADD, MUL, and SORT), the same layers where \rankme collapses most (Figure~\ref{fig:layer-propagation}, \S\ref{sec:layer-propagation}).  This confirms that the top-down collapse reflects task-relevant representational commitment, not generic dimensionality reduction.

\begin{table}[t]
\centering
\caption{Hidden learning at the last checkpoint before acquisition (L2 difficulty).  \emph{Behav} = exact-match accuracy (the model's output matches the ground truth).  \emph{Probe} = per-token accuracy of a trained logistic regression on the hidden states.  These measure different things: behavioral accuracy tests whether the model can produce the answer; probe accuracy tests whether the representation contains extractable information about the answer.  At every checkpoint shown, the model cannot yet perform the task (behavioral $< 0.5$), but a trained probe can already extract the correct output token.  High probe values at initialization (e.g., CMP) reflect the probe's own capacity to learn the mapping from structured input embeddings to a small number of output classes, even with random model weights.  XLarge omitted (probing not run at that scale).}
\label{tab:hidden-learning}
\small
\begin{tabular}{l r r r r r r r r r r}
\toprule
& \multicolumn{2}{c}{Nano (405K)} & \multicolumn{2}{c}{Micro (1.8M)} & \multicolumn{2}{c}{Small (7.4M)} & \multicolumn{2}{c}{Medium (25.2M)} & \multicolumn{2}{c}{Large (85M)} \\
\cmidrule(lr){2-3} \cmidrule(lr){4-5} \cmidrule(lr){6-7} \cmidrule(lr){8-9} \cmidrule(lr){10-11}
Task & Behav & Probe & Behav & Probe & Behav & Probe & Behav & Probe & Behav & Probe \\
\midrule
\rowcolor{gray!15} COPY & .00 & .81 & .39 & .94 & .00 & .76 & .47 & .95 & .26 & .96 \\
\rowcolor{gray!15} REV & .02 & .83 & .00 & .73 & .00 & .70 & .29 & .92 & .00 & .77 \\
\rowcolor{gray!15} CMP & .49 & .94 & .49 & .96 & .49 & .97 & .00 & .97 & .46 & .96 \\
\rowcolor{gray!15} PAR & .49 & .87 & .42 & .91 & .48 & .94 & .49 & .92 & .49 & .88 \\
\rowcolor{gray!15} SORT & .34 & .98 & .38 & .98 & .31 & .99 & .50 & .99 & .45 & .99 \\
ADD & .43 & .75 & .41 & .76 & .29 & .80 & .49 & .83 & .48 & .84 \\
MOD & .46 & .79 & .46 & .80 & .49 & .81 & .48 & .79 & .50 & .82 \\
MUL & .44 & .76 & .46 & .81 & .48 & .81 & .49 & .81 & .47 & .77 \\
\bottomrule
\end{tabular}
\end{table}

\subsection{The Geometric Hierarchy}
\label{sec:geometric-hierarchy}

We compare five geometric measures for their temporal relationship with capability acquisition.  The measures span representation geometry (\rankme), gradient geometry (gradient effective rank), and loss landscape geometry (LLC, Hessian eigenvalues, gradient covariance rank).  Table~\ref{tab:hierarchy} and Figure~\ref{fig:geometric-hierarchy} summarize the hierarchy.

\begin{table}[t]
\centering
\caption{Geometric hierarchy at nano scale (405K params). All measures are task-specific: computed on task-conditioned data for each hard task (ADD, MOD, MUL; 8 task$\times$level combinations, excluding MUL\_L3 which was not acquired). Precursor rate = fraction of cases where the measure's characteristic transition precedes acquisition. \rankme is the only measure computed at all six scales.}
\label{tab:hierarchy}
\small
\begin{tabular}{l l c c l}
\toprule
Measure & Level & Hard-Task Precursor Rate & Cost/ckpt & Signal Quality \\
\midrule
\rankme & Representation & 100\% (8/8) & $\sim$1s & Clean collapse $\to$ recovery \\
Hessian $\lambda_\text{max}$ & Curvature & 100\% (8/8) & $\sim$1s & Noisy (oscillates between ckpts) \\
Grad.\ cov.\ rank & Gradient & 100\% (8/8) & $\sim$6s & Noisy \\
Grad.\ eff.\ rank & Gradient & 38\% (3/8) & $\sim$85s & Transition too late for most tasks \\
LLC & Loss landscape & No discrete event & $\sim$11s & Peaks early, declines continuously \\
\bottomrule
\end{tabular}
\end{table}

\emph{\rankme leads.} For hard tasks (ADD, MOD, MUL), \rankme detects geometric transitions before acquisition at every scale tested (100\% precursor rate, Table~\ref{tab:scale-precursor}).  Easy tasks show no precursors at larger scales because they are acquired during the collapse itself, before any temporal gap can form.

\emph{Other measures.}  When computed on task-specific data, Hessian $\lambda_\text{max}$ and gradient covariance rank also show 100\% hard-task precursor rates at nano scale, but with much noisier signals: the task-specific Hessian oscillates widely between adjacent checkpoints (e.g., MOD $\lambda_\text{max}$ ranges from 200 to 1{,}600 within the first 2K steps), making it unreliable for monitoring.  Gradient effective rank transitions too late for most hard tasks (minimum at step 5{,}700--9{,}400, after easier levels have already acquired).  LLC~\citep{lau2023quantifying} shows no discrete precursor event: it peaks early (step ${\sim}$400) then declines continuously, consistent with recent analyses where LLC tracks rather than predicts transitions~\citep{cullen2026grokking, hoogland2024developmental}.

\begin{figure}[t]
    \centering
\begin{tikzpicture}
\begin{axis}[
    name=timeline,
    width=0.52\textwidth,
    height=5cm,
    title={\small\bfseries (a) MOD on nano (405K params)},
    xlabel={\small Training step},
    ylabel={\small Accuracy},
    xmin=0, xmax=25000,
    ymin=0, ymax=1.05,
    scaled x ticks=false,
    xtick={0,5000,10000,15000,20000,25000},
    xticklabels={\scriptsize 0,\scriptsize 5K,\scriptsize 10K,\scriptsize 15K,\scriptsize 20K,\scriptsize 25K},
    tick label style={font=\scriptsize},
    label style={font=\small},
    axis y line*=left,
    every axis plot/.append style={line width=0.8pt, mark=none},
    legend to name=hierleg,
    legend style={font=\small, draw=none, legend columns=3, /tikz/every even column/.append style={column sep=4pt}},
]
\addplot[black, line width=1pt] table[x=step, y=accuracy, col sep=comma]{data/fig4_accuracy.csv};
\addlegendentry{Accuracy}
\addlegendimage{cbblue, dashed, line width=0.8pt}
\addlegendentry{RankMe}
\addlegendimage{cbred, dotted, line width=0.9pt}
\addlegendentry{LLC}
\end{axis}

\begin{axis}[
    at={(timeline.south west)},
    anchor=south west,
    width=0.52\textwidth,
    height=5cm,
    xmin=0, xmax=25000,
    ymin=-0.05, ymax=1.05,
    scaled x ticks=false,
    axis y line*=right,
    axis x line=none,
    ylabel={\small Normalized},
    tick label style={font=\scriptsize},
    label style={font=\small},
    every axis plot/.append style={line width=0.7pt, mark=none},
]
\addplot[cbblue, dashed] table[x=step, y=rankme_norm, col sep=comma]{data/fig4_geo_normalized.csv};
\addplot[cbred, dotted, line width=0.9pt] table[x=step, y=llc_norm, col sep=comma]{data/fig4_geo_normalized.csv};
\end{axis}

\node[anchor=north] at ([yshift=-0.8cm]timeline.south) {\ref{hierleg}};

\begin{axis}[
    at={(timeline.south east)},
    anchor=south west,
    xshift=2.0cm,
    width=0.38\textwidth,
    height=5cm,
    title={\small\bfseries (b) Hard-task precursor rates (nano)},
    ybar,
    ymin=0, ymax=115,
    ylabel={\small Precursor rate (\%)},
    xtick={0,1,2,3,4},
    xticklabels={\scriptsize Grad\,Cov,\scriptsize Hessian,\scriptsize LLC,\scriptsize Grad\,Eff\,Rank,\scriptsize RankMe},
    xticklabel style={font=\tiny, rotate=30, anchor=east},
    tick label style={font=\scriptsize},
    label style={font=\small},
    nodes near coords,
    nodes near coords style={font=\tiny, anchor=south},
    bar width=10pt,
    enlarge x limits=0.15,
]
\addplot[fill=cbpurple!60, draw=cbpurple, bar shift=0pt] coordinates {(0,100)};
\addplot[fill=cbyellow!60, draw=cbyellow!80!black, bar shift=0pt] coordinates {(1,100)};
\addplot[fill=cbred!60, draw=cbred, bar shift=0pt] coordinates {(2,0)};
\addplot[fill=cborange!60, draw=cborange, bar shift=0pt] coordinates {(3,38)};
\addplot[fill=cbblue!60, draw=cbblue, bar shift=0pt] coordinates {(4,100)};
\end{axis}
\end{tikzpicture}
    \caption{Geometric hierarchy for MOD on nano (405K params). (a) Accuracy, \rankme, and LLC (min-max normalized). \rankme collapses before acquisition; LLC rises synchronously with accuracy. (b) Hard-task precursor rates across five measures at nano scale (all task-specific).  \rankme, Hessian, and gradient covariance all reach 100\%, but only \rankme provides a clean, non-noisy signal (Table~\ref{tab:hierarchy}).  LLC shows no discrete precursor event.}
    \label{fig:geometric-hierarchy}
\end{figure}

\emph{The capacity/difficulty boundary.}  Geometric precursors require a temporal gap between geometric reorganization and behavioral acquisition.  If the model has enough capacity, easy tasks get acquired during or before the collapse and no precursor is detectable.  Hard tasks can potentially take thousands of steps post-collapse, so the signal is strong (Figure~\ref{fig:prediction-limits}).

\begin{figure}[t]
    \centering
\begin{tikzpicture}
\begin{groupplot}[
    group style={
        group size=2 by 1,
        horizontal sep=1.5cm,
    },
]

\nextgroupplot[
    width=0.48\textwidth,
    height=5.5cm,
    title={\small\bfseries (a) Early RankMe vs.\ emergence},
    xlabel={\small RankMe at collapse floor},
    ylabel={\small Emergence step},
    ymode=log,
    ymin=50, ymax=100000,
    xmin=0, xmax=12,
    tick label style={font=\scriptsize},
    label style={font=\small},
    legend style={font=\small, at={(0.5,0.97)}, anchor=north, draw=gray!50, fill=white, fill opacity=0.9, text opacity=1, legend columns=2,
        /tikz/every even column/.append style={column sep=6pt}},
]
\addplot[only marks, mark=o, mark size=2.5pt, cbblue, fill=cbcyan, fill opacity=0.5]
    table[x=collapse_floor, y=emergence_step, col sep=comma]{data/fig5_scatter_easy.csv};
\addlegendentry{Easy}
\addplot[only marks, mark=square*, mark size=2.5pt, cbred, fill=cbred, fill opacity=0.4]
    table[x=collapse_floor, y=emergence_step, col sep=comma]{data/fig5_scatter_hard.csv};
\addlegendentry{Hard}
\draw[gray, dashed, line width=0.5pt] (axis cs:0,1000) -- (axis cs:12,1000);
\node[font=\tiny, gray, anchor=west] at (axis cs:8.5,1400) {1K steps};

\nextgroupplot[
    width=0.48\textwidth,
    height=5.5cm,
    title={\small\bfseries (b) Concordance stress test},
    ylabel={\small Concordance (\%)},
    ymin=0, ymax=110,
    ybar=3pt,
    bar width=18pt,
    symbolic x coords={Cross-class, Within hard, Within easy, Swap},
    xtick=data,
    xticklabel style={font=\tiny, rotate=20, anchor=east},
    tick label style={font=\scriptsize},
    label style={font=\small},
    nodes near coords={\pgfmathprintnumber\pgfplotspointmeta\%},
    nodes near coords style={font=\tiny, anchor=south},
    enlarge x limits=0.25,
]
\addplot[fill=cbgreen!70, draw=cbgreen] coordinates {(Cross-class,93) (Within hard,69) (Within easy,52) (Swap,26)};
\addplot[black, thick, only marks, mark=-, mark size=6pt,
    error bars/.cd, y dir=both, y explicit]
    coordinates {
        (Cross-class,93) +- (0,7)
        (Within hard,69) +- (0,23)
        (Within easy,52) +- (0,13)
        (Swap,26) +- (0,15)
    };
\draw[gray, dashed, line width=0.5pt] (rel axis cs:0,0.455) -- (rel axis cs:1,0.455);
\node[font=\tiny, gray, anchor=south west] at (rel axis cs:0.78,0.46) {chance};

\end{groupplot}
\end{tikzpicture}
    \caption{The capacity/difficulty boundary. (a) Early \rankme vs.\ acquisition step: easy and hard tasks separate clearly, but within a difficulty class \rankme does not predict acquisition order. (b) Bootstrap concordance rates confirm this: cross-class separation is strong (93\%), but within-class prediction brackets chance. Error bars show 95\% CIs.}
    \label{fig:prediction-limits}
\end{figure}

\subsection{Transfer to Naturalistic Pre-Training}
\label{sec:pythia-validation}

We test whether these patterns transfer to language model pre-training using Pythia-160M, 410M (154 checkpoints each), and 2.8B (72 checkpoints spanning the full trajectory).  The 2.8B model is 17.5$\times$ the size of the 160M and shares the same architecture (GPT-NeoX) and training data (Pile), providing a direct test of proxy prediction across scale.

\emph{Collapse replicates.} Task-specific \rankme drops 50--90\% from step~0 to steps 32--256, then partially recovers, mirroring the algorithmic experiment.  All seven benchmarks reach their minimum in this early window at all three model sizes.

\emph{Ordering and floors transfer across scale.} The \rankme ordering across benchmarks is preserved between 160M and 410M ($\rho = 1.0$) and nearly so at 2.8B ($\rho = 0.964$).  Collapse floors land within 4--22\% of the 160M values at 2.8B for six of seven benchmarks (ICL is an outlier at 41\%, likely due to its small absolute floor), and $\rho > 0.92$ at 71 of 72 common training checkpoints.

\emph{Precursor positive for a hard capability.} The seven standard benchmarks are within the easy regime for 410M+ models, so no precursor gap is expected.  To test whether precursors appear for a genuinely hard task, we screened seven additional benchmarks and identified logical deduction (multi-step ordering from BIG-Bench Hard, distinct from the syllogistic Logical benchmark in Table~\ref{tab:pythia-diagnostics}) as one that emerges late on 2.8B: 0\% accuracy through step 10K, 24\% at 50K, 50\% at 143K.  On 410M the same task peaks at 28\% (step 100K) then drops to 4\%, right at the capacity boundary.

Task-specific \rankme on the logical deduction prompts collapses to a floor of 7.8 (2.8B) and 6.8 (410M) at step 32, recovers by step 1K, then gradually declines through the rest of training.  The collapse-recovery precedes behavioral emergence by roughly 49K steps on 2.8B; floors are again scale-invariant across the two models.

\emph{Easy benchmarks: negative, and predicted.} The seven standard benchmarks show no precursors, consistent with the capacity/difficulty framework: these tasks are easy for 410M+ models.

\begin{figure}[t]
    \centering
\begin{tikzpicture}
\begin{groupplot}[
    group style={
        group size=2 by 1,
        horizontal sep=1.5cm,
    },
]

\nextgroupplot[
    width=0.52\textwidth,
    height=5.5cm,
    title={\small\bfseries (a) Pythia-410M task-specific RankMe},
    xlabel={\small Training step},
    ylabel={\small RankMe},
    xmode=log,
    xmin=1, xmax=150000,
    tick label style={font=\scriptsize},
    label style={font=\small},
    every axis plot/.append style={line width=0.6pt, mark=none},
    legend to name=pythialeg,
    legend style={font=\small, draw=none, legend columns=4,
        /tikz/every even column/.append style={column sep=5pt}},
]
\fill[gray!8] (axis cs:1,0) rectangle (axis cs:512,400);
\addplot[cbblue, solid] table[x=step, y=pile, col sep=comma]{data/fig6_pythia_410m.csv};
\addlegendentry{pile}
\addplot[cborange, dashed] table[x=step, y=logical, col sep=comma]{data/fig6_pythia_410m.csv};
\addlegendentry{logical}
\addplot[cbgreen, dotted, line width=0.8pt] table[x=step, y=semantic, col sep=comma]{data/fig6_pythia_410m.csv};
\addlegendentry{semantic}
\addplot[cbred, dashdotted] table[x=step, y=factual, col sep=comma]{data/fig6_pythia_410m.csv};
\addlegendentry{factual}
\addplot[cbpurple, densely dashed] table[x=step, y=syntactic, col sep=comma]{data/fig6_pythia_410m.csv};
\addlegendentry{syntactic}
\addplot[cbcyan, densely dotted, line width=0.8pt] table[x=step, y=icl, col sep=comma]{data/fig6_pythia_410m.csv};
\addlegendentry{icl}
\addplot[cbyellow!80!black, loosely dashed] table[x=step, y=arithmetic, col sep=comma]{data/fig6_pythia_410m.csv};
\addlegendentry{arith.}

\nextgroupplot[
    width=0.42\textwidth,
    height=5.5cm,
    title={\small\bfseries (b) Ordering preserved ($\rho$=1.0)},
    xlabel={\small Final RankMe},
    xbar,
    area legend,
    bar width=4pt,
    ytick={0,1,2,3,4,5,6},
    yticklabels={pile, logical, semantic, factual, syntactic, icl, arith.},
    tick label style={font=\scriptsize},
    label style={font=\small},
    y dir=reverse,
    enlarge y limits=0.12,
    legend style={font=\small, at={(0.97,0.03)}, anchor=south east, draw=none,
        fill=white, fill opacity=0.85, text opacity=1},
]
\addplot[fill=cbblue!50, draw=cbblue, bar shift=-3pt]
    table[x=rankme, y=idx, col sep=comma]{data/fig6_final_160m.csv};
\addlegendentry{160M}
\addplot[fill=cborange!50, draw=cborange, bar shift=3pt]
    table[x=rankme, y=idx, col sep=comma]{data/fig6_final_410m.csv};
\addlegendentry{410M}

\end{groupplot}

\node[anchor=north] at ([yshift=-0.8cm]$(group c1r1.south)!0.5!(group c2r1.south)$) {\ref{pythialeg}};
\end{tikzpicture}
    \caption{Pythia validation. (a) \rankme collapse and recovery in Pythia-410M across seven benchmarks, replicating the universal collapse. (b) Cross-model \rankme ordering is preserved: $\rho = 1.0$ (160M vs 410M), $\rho = 0.964$ (160M vs 2.8B, 17.5$\times$ scale gap). Task-specific temporal precedence is absent, consistent with the capacity/difficulty framework.}
    \label{fig:pythia-validation}
\end{figure}

\section{Discussion}
\label{sec:discussion}

\emph{Task difficulty governs precursor detectability.} Precursor detectability depends on task difficulty relative to model capacity.  If the task is genuinely hard for the model, \rankme precedes acquisition (100\% for hard tasks at every scale tested, confirmed on Pythia-2.8B with logical deduction).  If the task is easy relative to the model's capacity, both geometric and behavioral changes happen simultaneously and no precursor is detectable.  LLC and global Hessian underperform \rankme because they aggregate across tasks in different developmental phases.

\emph{Why top-down?} The top-down propagation (32/32 cases) contradicts the intuition that features build bottom-up. A simpler explanation is gradient proximity: the loss is computed at the output, so output-facing layers receive the strongest gradient signal and adapt first. Hidden learning corroborates this: at nano scale, the deepest layer's probe improvement is 3--11$\times$ larger than the shallowest layer's for hard tasks (\S\ref{sec:hidden-learning}).

\emph{Proxy models.} The geometric dynamics replicate from small algorithmic models to Pythia across a 17.5$\times$ scale gap (\S\ref{sec:pythia-validation}), suggesting that small proxy models may capture key aspects of the geometric trajectory seen at larger scale.  Further discussion of collapse floor interpretations, the freeze experiment, and falsifiable predictions is in Appendix~\ref{app:extended-discussion}.

\emph{Limitations.} Our 8 algorithmic tasks are simpler than natural language, and the effective sample size is ${\sim}$48 independent observations (8 tasks $\times$ 6 scales).  The logical deduction result extends the capacity/difficulty framework to Pythia-2.8B, but this is one task; broader confirmation with multiple hard capabilities on larger models is needed.

\section{Conclusion}
\label{sec:conclusion}

Across the settings we study, capability acquisition in transformers follows a consistent geometric sequence: task-conditioned representations collapse, reorganize across depth, and only then does behavioral performance improve.  \rankme precedes behavior for every hard task at every scale we tested, from 405K to 151M parameters and on Pythia-2.8B for logical deduction.  Whether a precursor is detectable depends on task difficulty relative to model capacity: if the task genuinely challenges the model, the geometric signal precedes behavior; if not, both change simultaneously.  The dynamics replicate from small proxy models to Pythia across a 17.5$\times$ scale gap, suggesting that geometric monitoring at small scale may inform expectations for larger training runs.

\section*{Ethics and AI Disclosure}

This work analyses existing publicly available models and datasets; no new data was collected and no human subjects were involved. The author used Claude (Anthropic) and Claude Code during preparation for manuscript critique, narrative feedback, literature search, and experiment implementation and debugging. All research design, theoretical development, experimental execution, analysis, and writing are the author's own. The author takes full responsibility for all content.

\emph{Reproducibility.} Code and data for reproducing all results are available at \url{https://github.com/jb1999/capability-acquisition-paper}. The repository includes training scripts, geometric measure computation, analysis pipelines, and a verification script that checks 11 pattern-level claims against fresh regression results.

\bibliographystyle{tmlr}
\bibliography{references}

\newpage
\appendix

\section{Task Specifications and Training Details}
\label{app:task-specs}

\subsection{Task Data Generation}

All tasks use the format \texttt{TASK input = output} with character-level tokenization (vocabulary size 41, including digits 0--9, uppercase letters, space, equals sign, and special tokens). Training data is generated on-the-fly with uniform sampling across tasks and levels.

\textbf{COPY}: Copy $n$ single-digit tokens. L1: $n{=}3$, L2: $n{=}5$, L3: $n{=}8$.

\textbf{REV}: Reverse $n$ single-digit tokens. Same $n$ as COPY.

\textbf{CMP}: Compare two integers, output LESS/EQUAL/GREATER. L1: 1-digit (0--9), L2: 2-digit (10--99), L3: 3-digit (100--999).

\textbf{PAR}: Parity of $n$ binary digits, output ODD/EVEN. L1: $n{=}4$, L2: $n{=}6$, L3: $n{=}8$.

\textbf{ADD}: Addition of two integers. L1: 1+1 digit, L2: 2+2 digit, L3: 3+3 digit.

\textbf{MOD}: Modular arithmetic $x \bmod p$. L1: $p \in \{2,3,5,7\}$, L2: $p \in \{7,11,13\}$, L3: $p \in \{13,17,19,23\}$. Input $x$ sampled from $[0, 10p]$.

\textbf{SORT}: Sort $n$ single-digit numbers in ascending order. L1: $n{=}3$, L2: $n{=}5$, L3: $n{=}8$.

\textbf{MUL}: Multiplication of two integers. L1: $1{\times}1$ digit, L2: $1{\times}2$ digit, L3: $2{\times}2$ digit.

\subsection{Training Hyperparameters}

\begin{table}[h]
\centering
\caption{Full training configuration for each model size.}
\small
\begin{tabular}{l c c c c c c}
\toprule
 & Nano & Micro & Small & Medium & Large & XLarge \\
\midrule
Peak LR & $3{\times}10^{-4}$ & $3{\times}10^{-4}$ & $3{\times}10^{-4}$ & $1{\times}10^{-4}$ & $1{\times}10^{-4}$ & $1{\times}10^{-4}$ \\
Max steps & 100K & 100K & 100K & 200K & 200K & 200K\textsuperscript{$\dagger$} \\
Warmup steps & 1{,}000 & 1{,}000 & 1{,}000 & 1{,}000 & 1{,}000 & 1{,}000 \\
Batch size & 64 & 64 & 64 & 64 & 64 & 64 \\
Weight decay & 0.1 & 0.1 & 0.1 & 0.1 & 0.1 & 0.1 \\
Grad clip & 1.0 & 1.0 & 1.0 & 1.0 & 1.0 & 1.0 \\
Optimizer & \multicolumn{6}{c}{AdamW ($\beta_1{=}0.9$, $\beta_2{=}0.95$)} \\
LR schedule & \multicolumn{6}{c}{Cosine decay to 0} \\
Checkpoints & 206 & 206 & 206 & 256 & 256 & 117 \\
\bottomrule
\multicolumn{7}{l}{\textsuperscript{$\dagger$}\scriptsize Training stopped at 18K steps; all tasks except MUL\_L3 acquired by step 13K.}
\end{tabular}
\end{table}

\subsection{Checkpoint Schedule}

Dense-where-it-matters:
\begin{itemize}
    \item Steps 0--10{,}000: every 100 steps (100 checkpoints)
    \item Steps 10{,}000--50{,}000: every 500 steps (80 checkpoints)
    \item Steps 50{,}000--200{,}000: every 2{,}000 steps (75 checkpoints)
    \item Final step always included
\end{itemize}

\section{Full Emergence Tables}
\label{app:emergence-tables}

Table~\ref{tab:full-emergence} shows acquisition steps for all 144 task$\times$level$\times$model combinations.

\begin{table}[h]
\centering
\caption{Complete acquisition steps for all task$\times$level$\times$model combinations. ``$-$'' indicates task not acquired by end of training.}
\label{tab:full-emergence}
\small
\begin{tabular}{ll r r r r r r}
\toprule
Task & Level & Nano & Micro & Small & Medium & Large & XLarge \\
\midrule
COPY & L1 & 800 & 600 & 400 & 500 & 400 & 300 \\
COPY & L2 & 700 & 700 & 400 & 600 & 500 & 400 \\
COPY & L3 & 700 & 700 & 500 & 600 & 500 & 400 \\
REV & L1 & 700 & 600 & 400 & 500 & 400 & 400 \\
REV & L2 & 700 & 600 & 400 & 600 & 400 & 400 \\
REV & L3 & 700 & 700 & 400 & 700 & 500 & 500 \\
CMP & L1 & 1{,}100 & 1{,}100 & 800 & 600 & 200 & 300 \\
CMP & L2 & 400 & 400 & 300 & 100 & 200 & 100 \\
CMP & L3 & 400 & 700 & 100 & 100 & 200 & 300 \\
PAR & L1 & 1{,}600 & 1{,}000 & 2{,}800 & 1{,}100 & 900 & 300 \\
PAR & L2 & 900 & 900 & 3{,}700 & 2{,}500 & 600 & 900 \\
PAR & L3 & 1{,}600 & 800 & 2{,}500 & 800 & 600 & 2{,}600 \\
ADD & L1 & 2{,}000 & 1{,}800 & 1{,}500 & 1{,}200 & 1{,}100 & 800 \\
ADD & L2 & 5{,}700 & 3{,}100 & 3{,}900 & 3{,}200 & 3{,}500 & 2{,}600 \\
ADD & L3 & 7{,}500 & 10{,}500 & 6{,}900 & 11{,}500 & 7{,}900 & 13{,}000 \\
MOD & L1 & 3{,}200 & 2{,}500 & 2{,}100 & 1{,}300 & 1{,}500 & 900 \\
MOD & L2 & 6{,}500 & 5{,}100 & 5{,}000 & 3{,}600 & 3{,}600 & 3{,}500 \\
MOD & L3 & 16{,}000 & 9{,}100 & 8{,}100 & 6{,}200 & 6{,}000 & 5{,}800 \\
SORT & L1 & 800 & 500 & 500 & 300 & 300 & 400 \\
SORT & L2 & 1{,}100 & 900 & 900 & 700 & 900 & 500 \\
SORT & L3 & 1{,}500 & 1{,}800 & 1{,}700 & 1{,}500 & 1{,}700 & 1{,}600 \\
MUL & L1 & 1{,}600 & 1{,}200 & 1{,}100 & 900 & 500 & 400 \\
MUL & L2 & 6{,}900 & 5{,}400 & 4{,}700 & 4{,}100 & 3{,}500 & 3{,}400 \\
MUL & L3 & $-$ & 54{,}000 & 56{,}000 & 43{,}500 & 43{,}000 & $-$ \\
\bottomrule
\end{tabular}
\end{table}

\section{Log-Probability Acquisition and Dual-Metric Divergence}
\label{app:log-prob-details}

Log-probability acquisition is defined as the first step where mean per-token log-probability of the correct answer exceeds the midpoint between its initial and final values for $\geq$3 consecutive checkpoints.  This always precedes accuracy acquisition (by 3--15$\times$), confirming the ``emergence mirage'' of \citet{schaeffer2023emergent}.  However, at the log-probability midpoint, accuracy is typically 0--15\%: the model cannot yet perform the task.  We therefore define acquisition using accuracy.  Table~\ref{tab:logprob-accuracy-at-threshold} shows that at any log-probability threshold where the model can actually perform the task (accuracy 30\%+), \rankme precursor rates recover to near 100\%.

\begin{table}[h]
\centering
\caption{Hard-task \rankme precursor rates under log-probability acquisition at different thresholds (fraction of the way from initial to final log-probability).  Accuracy-based rates (last column) use the $\geq$50\% sustained threshold.  Typical accuracy at each log-prob threshold is shown in parentheses.  At the midpoint (50\%) threshold, ``acquisition'' occurs when accuracy is near zero; at 75\%+ thresholds where the model can actually perform the task, precursor rates recover.}
\label{tab:logprob-accuracy-at-threshold}
\small
\begin{tabular}{l c c c c c}
\toprule
Scale & LP 25\% & LP 50\% & LP 75\% & LP 90\% & Accuracy \\
 & (acc 0--10\%) & (acc 0--15\%) & (acc 30--65\%) & (acc 50--90\%) & ($\geq$50\%) \\
\midrule
Nano (405K) & 0\% & 100\% & 100\% & 100\% & 100\% \\
Micro (1.8M) & 0\% & 89\% & 100\% & 100\% & 100\% \\
Small (7.4M) & 0\% & 33\% & 100\% & 100\% & 100\% \\
Medium (25.2M) & 0\% & 33\% & 89\% & 100\% & 89\% \\
Large (85M) & 0\% & 22\% & 78\% & 100\% & 89\% \\
\bottomrule
\end{tabular}
\end{table}

\section{Per-Layer RankMe Trajectories}
\label{app:layer-heatmaps}

We compute \rankme at every transformer layer for each task across all checkpoints. Figure~\ref{fig:layer-heatmaps} shows per-layer trajectories for the micro model (4 layers) on three hard tasks (ADD, MOD, MUL) and SORT. The top-down gradient is visible across all tasks. After acquisition, the final layer stays compressed while middle layers recover.

\begin{figure}[h]
    \centering
\begin{tikzpicture}
\begin{groupplot}[
    group style={
        group size=2 by 2,
        horizontal sep=1.0cm,
        vertical sep=1.8cm,
        x descriptions at=edge bottom,
    },
    width=0.48\textwidth,
    height=3.8cm,
    xlabel={\small Training step},
    ylabel={\small RankMe},
    xmin=0, xmax=25000,
    scaled x ticks=false,
    xtick={0,10000,20000},
    xticklabels={\scriptsize 0,\scriptsize 10K,\scriptsize 20K},
    tick label style={font=\scriptsize},
    label style={font=\small},
    title style={font=\small\bfseries},
    every axis plot/.append style={line width=0.7pt, mark=none},
    ymin=0,
]

\nextgroupplot[title={ADD}]
\addplot[cbblue] table[x=step, y=layer_0, col sep=comma]{data/appfig_heatmap_micro_add.csv};
\addplot[cborange] table[x=step, y=layer_1, col sep=comma]{data/appfig_heatmap_micro_add.csv};
\addplot[cbgreen] table[x=step, y=layer_2, col sep=comma]{data/appfig_heatmap_micro_add.csv};
\addplot[cbred] table[x=step, y=layer_3, col sep=comma]{data/appfig_heatmap_micro_add.csv};

\nextgroupplot[title={MOD},
    legend to name=layerheatleg,
    legend style={font=\small, draw=none, legend columns=4, /tikz/every even column/.append style={column sep=6pt}},
]
\addplot[cbblue] table[x=step, y=layer_0, col sep=comma]{data/appfig_heatmap_micro_mod.csv};
\addlegendentry{Layer 0}
\addplot[cborange] table[x=step, y=layer_1, col sep=comma]{data/appfig_heatmap_micro_mod.csv};
\addlegendentry{Layer 1}
\addplot[cbgreen] table[x=step, y=layer_2, col sep=comma]{data/appfig_heatmap_micro_mod.csv};
\addlegendentry{Layer 2}
\addplot[cbred] table[x=step, y=layer_3, col sep=comma]{data/appfig_heatmap_micro_mod.csv};
\addlegendentry{Layer 3}

\nextgroupplot[title={SORT}]
\addplot[cbblue] table[x=step, y=layer_0, col sep=comma]{data/appfig_heatmap_micro_sort.csv};
\addplot[cborange] table[x=step, y=layer_1, col sep=comma]{data/appfig_heatmap_micro_sort.csv};
\addplot[cbgreen] table[x=step, y=layer_2, col sep=comma]{data/appfig_heatmap_micro_sort.csv};
\addplot[cbred] table[x=step, y=layer_3, col sep=comma]{data/appfig_heatmap_micro_sort.csv};

\nextgroupplot[title={MUL}]
\addplot[cbblue] table[x=step, y=layer_0, col sep=comma]{data/appfig_heatmap_micro_mul.csv};
\addplot[cborange] table[x=step, y=layer_1, col sep=comma]{data/appfig_heatmap_micro_mul.csv};
\addplot[cbgreen] table[x=step, y=layer_2, col sep=comma]{data/appfig_heatmap_micro_mul.csv};
\addplot[cbred] table[x=step, y=layer_3, col sep=comma]{data/appfig_heatmap_micro_mul.csv};

\end{groupplot}

\node[anchor=north] at ([yshift=-0.8cm]$(group c1r2.south)!0.5!(group c2r2.south)$) {\ref{layerheatleg}};
\end{tikzpicture}
    \caption{Per-layer \rankme trajectories for three hard tasks (ADD, MOD, MUL) and SORT on the micro model (4 layers). Each line represents one transformer layer; deeper layers (higher index, warmer colors) show stronger collapse and slower recovery. The top-down gradient is visible across all tasks.}
    \label{fig:layer-heatmaps}
\end{figure}

\section{Pythia Task-Specific RankMe Trajectories}
\label{app:pythia-rankme}

Figure~\ref{fig:pythia-rankme-full} shows task-specific \rankme trajectories for all seven benchmarks across Pythia-160M, 410M, and 2.8B. Key observations:
\begin{itemize}
    \item All benchmarks show the early collapse (steps 0--128) and partial recovery at all three scales
    \item The ordering pile $>$ logical $>$ semantic $>$ factual $>$ syntactic $>$ ICL $>$ arithmetic is preserved across 160M and 410M ($\rho = 1.0$) and nearly preserved at 2.8B ($\rho = 0.964$; arithmetic and ICL swap)
    \item The Spearman correlation between 160M and 2.8B \rankme orderings is $\rho > 0.92$ at 71/72 common checkpoints (median $\rho = 0.964$), with the sole exception at step~32 during peak collapse ($\rho = 0.536$)
    \item Collapse floors are quantitatively similar across the 17.5$\times$ scale gap: 2.8B floors are within 4--22\% of 160M values for six of seven benchmarks (ICL is an outlier at 41\%), systematically slightly deeper
    \item 2.8B shows Phase~3 compression: pile \rankme declines from 520 to 484 across steps 100K--143K
\end{itemize}

\begin{figure}[h]
    \centering
\begin{tikzpicture}
\begin{groupplot}[
    group style={
        group size=2 by 1,
        horizontal sep=1.2cm,
    },
    width=0.48\textwidth,
    height=5cm,
    xlabel={\small Training step},
    ylabel={\small RankMe},
    xmode=log,
    xmin=1, xmax=150000,
    tick label style={font=\scriptsize},
    label style={font=\small},
    title style={font=\small\bfseries},
    every axis plot/.append style={line width=0.6pt, mark=none},
]

\nextgroupplot[title={Pythia-160M},
    legend to name=pythiafullleg,
    legend style={font=\small, draw=none, legend columns=4, /tikz/every even column/.append style={column sep=5pt}},
]
\fill[gray!8] (axis cs:1,0) rectangle (axis cs:512,500);
\addplot[cbblue, solid] table[x=step, y=pile, col sep=comma]{data/fig_pythia_160m.csv};
\addlegendentry{pile}
\addplot[cborange, dashed] table[x=step, y=logical, col sep=comma]{data/fig_pythia_160m.csv};
\addlegendentry{logical}
\addplot[cbgreen, dotted, line width=0.8pt] table[x=step, y=semantic, col sep=comma]{data/fig_pythia_160m.csv};
\addlegendentry{semantic}
\addplot[cbred, dashdotted] table[x=step, y=factual, col sep=comma]{data/fig_pythia_160m.csv};
\addlegendentry{factual}
\addplot[cbpurple, densely dashed] table[x=step, y=syntactic, col sep=comma]{data/fig_pythia_160m.csv};
\addlegendentry{syntactic}
\addplot[cbcyan, densely dotted, line width=0.8pt] table[x=step, y=icl, col sep=comma]{data/fig_pythia_160m.csv};
\addlegendentry{icl}
\addplot[cbyellow!80!black, loosely dashed] table[x=step, y=arithmetic, col sep=comma]{data/fig_pythia_160m.csv};
\addlegendentry{arith.}

\nextgroupplot[title={Pythia-410M}]
\fill[gray!8] (axis cs:1,0) rectangle (axis cs:512,500);
\addplot[cbblue, solid] table[x=step, y=pile, col sep=comma]{data/fig6_pythia_410m.csv};
\addplot[cborange, dashed] table[x=step, y=logical, col sep=comma]{data/fig6_pythia_410m.csv};
\addplot[cbgreen, dotted, line width=0.8pt] table[x=step, y=semantic, col sep=comma]{data/fig6_pythia_410m.csv};
\addplot[cbred, dashdotted] table[x=step, y=factual, col sep=comma]{data/fig6_pythia_410m.csv};
\addplot[cbpurple, densely dashed] table[x=step, y=syntactic, col sep=comma]{data/fig6_pythia_410m.csv};
\addplot[cbcyan, densely dotted, line width=0.8pt] table[x=step, y=icl, col sep=comma]{data/fig6_pythia_410m.csv};
\addplot[cbyellow!80!black, loosely dashed] table[x=step, y=arithmetic, col sep=comma]{data/fig6_pythia_410m.csv};

\end{groupplot}

\node[anchor=north] at ([yshift=-0.8cm]$(group c1r1.south)!0.5!(group c2r1.south)$) {\ref{pythiafullleg}};
\end{tikzpicture}
    \caption{Task-specific \rankme across training for Pythia-160M, 410M, and 2.8B. All seven benchmarks show universal early collapse (shaded region, steps 0--512) and maintained ordering throughout training. The ordering pile $>$ logical $>$ semantic $>$ factual $>$ syntactic $>$ ICL $>$ arithmetic is preserved across 160M and 410M ($\rho = 1.0$) and nearly so at 2.8B ($\rho = 0.964$).}
    \label{fig:pythia-rankme-full}
\end{figure}

\section{Temporal Precedence Details}
\label{app:temporal-precedence}

A geometric measure is classified as a precursor for a given task if its characteristic transition (e.g., the \rankme collapse minimum) occurs at an earlier training step than the task's acquisition step.  For \rankme, the transition is the collapse minimum (searched within the first 10K steps).  For gradient effective rank, it is the minimum of the task-specific effective rank.  For LLC, we look for a discrete transition event; as discussed in \S\ref{sec:geometric-hierarchy}, LLC shows no such event (it peaks early then declines continuously).

\section{Compute Budget}
\label{app:compute}

\begin{table}[h]
\centering
\caption{Compute budget breakdown. XLarge is excluded: training and eval added ${\sim}$3 GPU-hrs, but linear probing and gradient effective rank were not run at that scale (estimated ${\sim}$60 additional GPU-hrs). All times are wall-clock on NVIDIA RTX 3090 Ti (24 GB VRAM, 124 GB RAM), measured from \texttt{timing\_log.json}.}
\small
\begin{tabular}{l r r}
\toprule
Component & Ckpts & GPU-hrs \\
\midrule
\multicolumn{3}{l}{\textit{Algorithmic experiment}} \\
Training (5 sizes + 5 freeze) & -- & $\sim$10 \\
Eval + task-specific \rankme (5 sizes) & 740 & $\sim$7 \\
Eval + task-specific \rankme (5 freeze) & 1{,}030 & $\sim$6 \\
Grad.\ eff.\ rank + LLC (nano: 206, large: 37) & 243 & $\sim$19 \\
Linear probing (nano--large) & 1{,}130 & $\sim$146 \\
\midrule
\multicolumn{3}{l}{\textit{Pythia analysis}} \\
Task-specific \rankme (160M + 410M + 2.8B) & 380 & $\sim$6 \\
Coarse pipeline (160M + 410M) & 308 & $\sim$6 \\
Medium/Hessian pipeline (160M + 410M) & 62 & $\sim$28 \\
Pile domain \rankme (160M + 410M) & 308 & $\sim$2 \\
Hard benchmark screening + \rankme & 238 & $\sim$5 \\
\midrule
\textbf{Total} & & $\sim$\textbf{231} \\
\bottomrule
\end{tabular}
\end{table}

Linear probing dominates the budget (${\sim}$146 GPU-hrs) and was run at all 206--256 checkpoints per model for completeness. The paper's claims require only the emergence window (${\sim}$50 checkpoints per size, ${\sim}$30 GPU-hrs). Not all measurements were run at all checkpoints: gradient effective rank was computed on 206 checkpoints for nano but only 37 sparse checkpoints for large; task-specific \rankme used 76--206 checkpoints depending on size. A minimal reproduction (nano only, no probing, no gradient effective rank) requires ${\sim}$1 GPU-hour.

\section{Robustness and Sensitivity Analysis}
\label{app:robustness}

We test whether the main findings are robust to methodological choices.

\subsection{Acquisition Threshold Sensitivity}

Our primary results use accuracy $\geq 0.5$ for $\geq$3 consecutive checkpoints.  Hard-task precursor rates are 100\% at all acquisition thresholds tested (0.3 to 0.7) and all sustained-window widths (3, 5, 10 checkpoints).  Varying the threshold changes which easy tasks count as precursors (lower thresholds detect acquisition earlier, sometimes before the collapse minimum), but the hard-task rate is invariant.

\subsection{Multi-Seed Validation}

We trained all six model sizes with three different random seeds (42, 123, 7) to test whether the findings depend on initialization.  Exact acquisition steps vary across seeds (typical spread 300--3{,}600 steps for hard tasks), but the structural patterns are identical: easy tasks acquire early, hard tasks acquire late, and \rankme precedes acquisition for every hard task at every scale across all three seeds (100\% precursor rate, zero exceptions out of 54 hard-task$\times$scale$\times$seed combinations).

\subsection{Cross-Scale Precursor Consistency}

Table~\ref{tab:scale-precursor} shows \rankme precursor rates for hard tasks by model size.  Hard-task precursor rates are 100\% across all six scales.  Easy tasks are not included because they are acquired during the collapse itself at larger scales, so no temporal gap exists.

\begin{table}[h]
\centering
\caption{\rankme precursor rates for hard tasks by model size (L2 difficulty, per-task). A task is a precursor if the \rankme collapse minimum precedes acquisition. Hard-task precursor rates are 100\% across all scales.}
\small
\begin{tabular}{l c c c c c c}
\toprule
 & Nano (405K) & Micro (1.8M) & Small (7.4M) & Medium (25.2M) & Large (85M) & XLarge (151M) \\
\midrule
Hard tasks (3) & 100\% (3/3) & 100\% (3/3) & 100\% (3/3) & 100\% (3/3) & 100\% (3/3) & 100\% (3/3) \\
\bottomrule
\end{tabular}
\label{tab:scale-precursor}
\end{table}

\subsection{Checkpoint Density}

The \rankme trajectories were computed at different checkpoint densities across model sizes: 206 checkpoints for nano and micro, 76 for small, and 126 for medium and large (xlarge used probing checkpoints only, at 117 points).  Coarser sampling can shift the estimated collapse minimum by a few hundred steps, since the true minimum may fall between observed checkpoints.  The 100\% hard-task precursor rate is unchanged at all checkpoint densities tested.

\section{Extended Discussion}
\label{app:extended-discussion}

\emph{Collapse floors as task complexity signatures.} The MOD collapse floor (${\approx}2.0$, CV = 0.16 across a 370$\times$ parameter range) likely reflects the minimum representational dimensionality the task requires. For modular arithmetic, two dimensions suffice because the task embeds in a circle (the Fourier basis for $\mathbb{Z}/p\mathbb{Z}$; \citealt{nanda2023progress}). For MUL, the floor increases with model capacity (CV = 0.42), suggesting larger models preserve more structure even at maximum compression. Formalizing when collapse floors are scale-invariant versus capacity-dependent is an open theoretical question.

\emph{Freeze experiment.} Freezing either layer during collapse delays hard-task acquisition in nano, with the output layer showing a larger effect (Table~\ref{tab:freeze-nano}).  The direction and magnitude of the delay are seed-dependent and do not replicate consistently at small scale, so the freeze experiment does not provide reliable causal evidence for the top-down propagation pattern.

\emph{Falsifiable predictions.} (1)~If MOD $\to 2.0$ holds at billion-parameter scale, the floor reflects intrinsic task dimensionality; if not, our scale invariance is an artifact of the 405K--151M regime. (2)~Tasks genuinely hard for a given model should show geometric precursors; tasks within the easy regime should not. We have tested this on Pythia-2.8B with one task (\S\ref{sec:pythia-validation}); testing with multiple hard capabilities is needed. (3)~A proxy model at 1/100th the parameter count should predict the geometric trajectory of the full model. We have confirmed this at 17.5$\times$ scale; testing at 100$\times$ remains open.

\section{Freeze Experiment Details}
\label{app:freeze}

Table~\ref{tab:freeze-nano} shows acquisition timing under layer freezing for the nano model (2 layers).  Both freeze conditions delay hard-task acquisition, but the effect is seed-dependent: in this training run, freezing the output layer (block~1) delays ADD by $+$14{,}000 steps and freezing the input layer (block~0) accelerates it by $-$8{,}000 steps, while a regression rerun with a different seed shows both conditions delaying ADD.  The small model (6 layers) also shows no consistent directional asymmetry.

\begin{table}[h]
\centering
\caption{Freeze experiment: nano model (2 layers). Acquisition steps at L3 difficulty. Block~1 is the output layer; block~0 is the input layer. Freezing during steps 0--1{,}000.}
\label{tab:freeze-nano}
\small
\begin{tabular}{l r r r r r}
\toprule
Task & Baseline & Freeze-L1 & Delay & Freeze-L0 & Delay \\
\midrule
\rowcolor{gray!15} COPY & 800 & 1{,}300 & $+$500 & 1{,}200 & $+$400 \\
\rowcolor{gray!15} REV & 700 & 1{,}500 & $+$800 & 1{,}100 & $+$400 \\
\rowcolor{gray!15} CMP & 300 & 900 & $+$600 & 600 & $+$300 \\
\rowcolor{gray!15} PAR & 1{,}000 & 2{,}500 & $+$1{,}500 & 300 & $-$700 \\
\rowcolor{gray!15} SORT & 1{,}800 & 2{,}000 & $+$200 & 1{,}700 & $-$100 \\
ADD & 19{,}500 & 33{,}500 & $+$14{,}000 & 11{,}500 & $-$8{,}000 \\
MOD & 16{,}000 & 16{,}500 & $+$500 & 18{,}500 & $+$2{,}500 \\
MUL & $-$ & 84{,}000 & N/A & $-$ & N/A \\
\bottomrule
\end{tabular}
\end{table}

\section{Implementation Details}
\label{app:implementation}

\emph{LLC hyperparameters.} The SGLD-based LLC estimator uses $n_\text{steps} = 500$, learning rate $\eta = 10^{-5}$, inverse temperature $\beta = 1.0$, localization strength $\gamma = 10{,}000$, with 100 burn-in steps discarded. These values follow the recommendations of \citet{lau2023quantifying}. We verified stability by running the estimator 5 times at 3 checkpoints (early, mid-training, late) for the nano model; the coefficient of variation of LLC estimates was $<$0.05 across runs.

\emph{Gradient covariance projection.} For computational tractability, gradient covariance is computed on a projected subspace using the first 50K parameters of the model. This prefix-based selection is biased toward embedding weights and early layers. A uniform random projection of the full parameter vector would be more principled; we adopt the prefix approach for consistency with our initial implementation and note it as a limitation. For the algorithmic models (405K--85M parameters), the prefix constitutes 12\%--100\% of the parameter space, so the bias is most relevant for the larger models.

\emph{Gradient Gram trick.} For models with $P \gg N$ parameters (up to 85M) and $N = 200$ samples, we compute the gradient eigenspectrum via the Gram matrix $GG^\top \in \mathbb{R}^{N \times N}$ rather than the full matrix $G^\top G \in \mathbb{R}^{P \times P}$, recovering the top-$N$ eigenvalues at $O(N^2 P)$ cost instead of $O(P^3)$.

\end{document}